%% file: icra2023BiancoOgnibene.tex
\documentclass[letterpaper, 10 pt, conference]{ieeeconf}  

\IEEEoverridecommandlockouts                              

\usepackage{graphicx} 
\usepackage{booktabs}
\usepackage{multirow}
\usepackage[table,xcdraw]{xcolor}
\usepackage{cite}

\title{\LARGE \bf
Robot Learning  Theory of Mind through Self-Observation: Exploiting the Intentions-Beliefs Synergy
}

\author{Francesca Bianco$^{1}$ and Dimitri Ognibene$^{2,1,*}$
\thanks{*This work was not supported by any organization}
\thanks{$^{1}$University of Essex, Colchester, UK}%
\thanks{$^{2}$Università degli Studi di Milano Bicocca, Milano, Italy}%
\thanks{$^{*}$email:dimitri.ognibene@unimib.it}%
}

\begin{document}

\maketitle
\thispagestyle{empty}
\pagestyle{empty}

\begin{abstract}

In complex environments, where the human sensory system
reaches its limits, our behaviour is strongly driven by our
beliefs about the state of the world around us.
Accessing others' beliefs, intentions, or mental states in
general, could thus allow for more effective social
interactions in natural contexts. Yet these variables are
not directly observable.
Theory of Mind (TOM), the ability to attribute to other
agents' beliefs,  intentions, or mental states in general, 
 is a crucial feature of human social interaction and has
become of interest to the robotics community. 
Recently, new models that are able to  learn  TOM have been
introduced.
In this paper, we show the synergy between learning to
predict low-level mental states, such as intentions and
goals, and attributing high-level ones, such as beliefs. 
Assuming that learning of beliefs can take place by
observing own decision and beliefs estimation processes in  partially observable
environments and 
using a simple feed-forward deep learning model, we show
that when learning to predict others' intentions and
actions, faster and more accurate predictions can be
acquired if beliefs attribution is learnt simultaneously
with action and intentions prediction. 
We show that the learning performance improves even when
observing agents with a different decision process and is
higher when observing beliefs-driven chunks of behaviour. 
We propose that our architectural approach can be relevant
for the design of future adaptive social robots that should
be able to autonomously understand and assist human
partners in novel natural environments and tasks.

\end{abstract}

\section{INTRODUCTION}

Due to recent technological developments,  the interactions between AI and humans have become pervasive and heterogeneous, extending from  voice assistants or recommender systems supporting the online experience of millions of users to autonomous driving cars. Principled models to represent the human collaborators' needs are being adopted \cite{rossi2017user} while
robotic perception in complex environments is becoming more flexible and adaptive \cite{van2021active,chaplot2020object, ammirato2017Active,ramakrishnan2019emergence}  even in social contexts, robot sensory limits are starting to be actively managed \cite{ognibene2013towards,demiris2003distributed}. 
However, robots and intelligent systems still have a limited understanding of how sensory limits affect human partners' behaviour and lead them to rely on internal beliefs about the state of the world.
This strongly impacts human-robot mutual understanding \cite{TANIGUCHI2016144} and calls for an effort to transfer the advance in robot perception management to methods to better cope with human collaborators' perceptual limits\cite{wang2020toward,fuchs2021gaze,hu2022toward}.

The possibility of introducing in robots and AI systems a Theory of Mind (TOM) \cite{sodian2016theory}, the ability to attribute to other agents' beliefs,  intentions, or mental states in general, has recently raised  hopes to further improve  robots' social skills \cite{JARAETTINGER2019105,bianco2019,ccelikok2019interactive, cuzzolin_morelli_cirstea_sahakian_2020}. 
While some studies have explored human partners' tendency to attribute mental states to robots \cite{10.1145/3477322.3477332,zhang2019theory,Qiaosi2021,banks2020theory}, the expected practical impact of TOM led to a diverse set of TOM implementations on robots. 
Several implementations relied on hardwired agents and task models that could be applied to infer mental states in  settings known at design time \cite{baker2011bayesian,devin2016implemented,LEMAIGNAN201745,vinanzi2019would,winfield2018experiments}. 
A step forward is presented in \cite{persiani2021} with an algorithm to understand unknown agents relying upon  Belief-Desire-Intention models of previously met agents.

Recently, following  \cite{rabinowitz2018machine} seminal work, several models have introduced deep learning based  TOM implementations  \cite{wen2019probabilistic,chandra2020stylepredict,Oguntola2021,aru2022mind, nguyen2022learning,Vadala2023}. 
This novel approach, learning both beliefs and intention attribution, should allow improved collaboration and adaptive human-robot collaboration in complex environments through a better understanding of humans' mental states. 
In this paper, we explore if the data-driven approach proposed in \cite{rabinowitz2018machine} and related works leads to improved predictions of the partner's intentions, which is often the mental state with the highest impact on the interaction performance. 
The prediction of  partners' intentions, even within a system producing prediction on several others' unobservable mental states, such as beliefs, will still rely only on the processing of observable behavioural inputs, aka state-action trajectories. 
In a purely supervised learning setting, such as that proposed in \cite{rabinowitz2018machine}, it is not immediate why performing an additional set of predictions, increasing the demands on the social perception system, should result in higher accuracy for the prediction of others’ intentions. 
This approach introduces additional complexity and noise that may hinder performance (see \cite{shah2019feasibility}). 
Moreover, deep learning models as those proposed in \cite{rabinowitz2018machine} are usually data hungry, which may further affect the value of the approach.
These factors may be some of the reasons for the long time required for the full development of TOM in infants \cite{sodian2016theory,bianco2019transferring}.

While all these considerations sound  technically valid, our results with simplified versions of the architecture proposed in \cite{rabinowitz2018machine}  show that the original  hypothesis may be true: learning is faster and more accurate if it takes place simultaneously  for the prediction of both intentions and beliefs together. 
Our results also show that the impact of  learning beliefs attribution on intention prediction is stronger in conditions of strong partial observability, e.g. when the observed agent does not still know where his target is. 
We found that when the system learns to predict intentions and beliefs at the same time it can better disambiguate and discard unrelated objects that are or have been in the sensory field of the observed agent.
This can be particularly relevant for assistive applications, especially those based on egovision that can monitor the sensory state of the partners \cite{damen2020epic}. 
Our hypothesis is that this is due to the regularization effect of multitask learning when tasks don't present  conflict but synergy \cite{langdon2022meta,thrun2012learning,crawshaw2020multi,ruder2017overview}. 
Indeed the observed accuracy gain decreases when the dataset size exceeds a certain threshold and makes regularization less helpful. On the other side, when very limited experience is available, the performance is slightly worse for joint prediction, maybe explaining the necessity for the  complex and multi-system developmental  process  that  seems to characterize  humans \cite{sodian2016theory}.

Another issue of the approach proposed in \cite{rabinowitz2018machine}  is the lack of training samples, as others' mental states are usually not available  through direct  behaviour observation or in datasets for training or prediction. We propose that self-observation, the observation of the agent's own behaviours and the internal decision and beliefs estimation processes can be used as the training signal. We tested that mental states prediction can then be generalized to different agents, even if  within certain limits. 

From a cognitive modelling perspective, the proposed architecture detaches from the motor simulation tradition common in robotics \cite{demiris2003distributed,wolpert2003unifying,ognibene2013towards,giese2015neural,dindo2011motor}, where the motor control system is used  for both action execution  and  perception. In our model, the motor control system is only the  source of the reference signal for the training of the social perception system. Among the others, this strategy has the advantage of reducing the strong interference  between action and observation that a motor simulation account of social perception could present \cite{zwickel2010interference,chaminade2005motor,friston2011action}. It could also reduce the computational demands and delays that could be involved in simulating others' actions while it is still able to exploit the observer's (motor or action) knowledge for social perception \cite{meltzoff2007like}. 
The presented architecture can be related to the \textit{associative hypothesis} of social perception \cite{james1890principles} and in particular, to  Hommel et al.'s\cite{hommel2001theory} interpretation, as described in \cite{csibra2007obsessed}, which has extended the associative view to intention interpretation and proposes that associations between behaviours (actions) and underlying intentions (effect) develop from an early age and can be used in later  in life as a means to infer and predict others’ intentions and behaviours. 
The associative account has also been previously related to others’ understanding by proposals of its involvement in mirror neurons development (e.g., \cite{heyes2010mirror}), as well as in sensorimotor matching for imitation \cite{decety2003self}. The architecture proposed in the present paper differs from these accounts as it focuses  on beliefs \cite{demiris2007prediction}. Specifically, It relies on associations between explicit belief representations learnt through self-experience and consequent behaviours to improve the prediction performance of others’ beliefs-driven behaviours. 

\section{METHODS}
\subsection{Architecture}
\begin{figure}[thpb]
      \centering
      \framebox{\parbox{3.3in}{
      \includegraphics[clip, trim=1cm 0.25cm .25cm 0.25cm,width=3.3in]{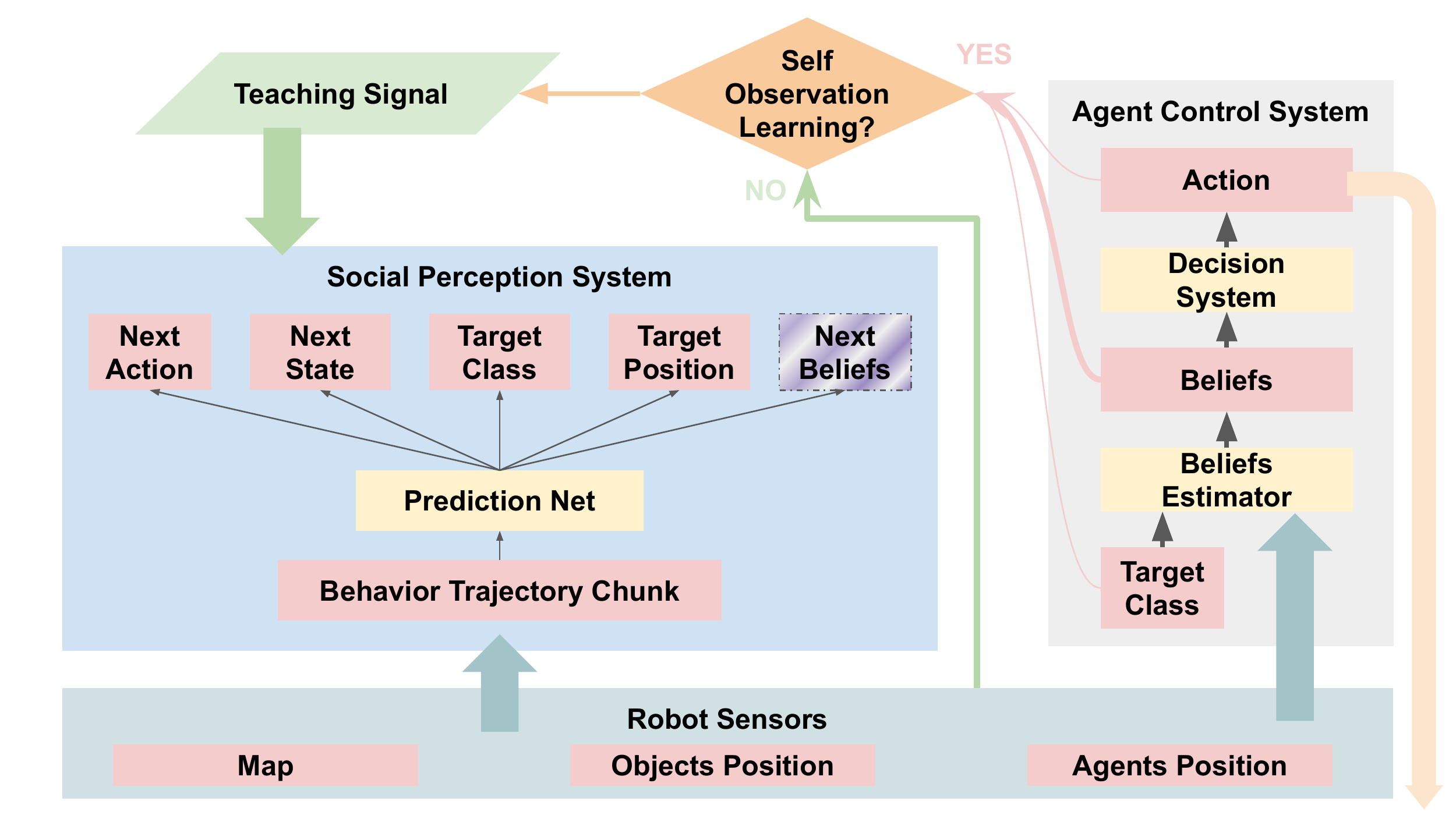}}}
      \caption{The architecture utilised in the here reported studies, formed of a shared prediction net torso and subsequently of separate prediction heads. For the NoBeliefs architecture, the following prediction heads are considered: 1. Target position, 2. Actor’s next action, and 3. Actor’s next state. For the Beliefs architecture, the 4. belief prediction head (in red) is also considered}
      \label{fig;architecture}
   \end{figure}

The agent is composed of two components (see Fig.\ref{fig;architecture}), one is the Social Perception System (SPS), that interpret and allows for predictions of observed agents behaviour, in terms of next actions, goals and even beliefs. The other component is the  control system that define the agent behaviour  to perform tasks similar to that it interprets in the SPS. The quantities, e.g. intentions, beliefs, actions, produced during the performance of the task are also used to fully train the SPS, while others' behaviours are used only for partial training, as others' beliefs cannot be observed.

\subsubsection{Social Perception System (SPS)}
The SPS’s goal is to make predictions about the observed actors’ future behaviour, with a specific interest on the actor's target position.
Two types of SPS were trained: \textbf{NoBelief SPS}, the baseline,  which predicts an actor’s target position, target class, next action, and next resulting state but not the beliefs; \textbf{Beliefs SPS}, was asked to predict, in addition to the previous, also the actor’s beliefs.

\paragraph{Input sensing and routing}
The SPS can observe themselves or others; the input vector for the system can then be provided by a common reference frame to represent either the self-localisation state of the observer, during self-observation learning, or the physical state of the other actor, both for learning and prediction of other actor's behaviour. 
Several architectures have studied the problem \cite{demiris2003distributed,wolpert2003unifying} of how to switch between the processing of own and others’ data and how to acquire others’ physical states. 
Note that in this case this function, while important, poses less constraints to behaviour performance as it feeds not the execution process but the social learning one \cite{chinellato2013time}. 

\paragraph{Input encoding and pre-processing}
The inputs is formed by a number (max 5 in the reported experiment) of past steps of a trajectory on a single grid map.
Observed actions-states pairs are combined through a spatialisation-concatenation operation, whereby actions are tiled over space into a tensor and concatenated to form a single tensor of shape (11 x 11 x 20).
While 11 x 11 represents the size of the grid world environments, 20 vectors are provided as inputs consisting of information regarding (a) actions (9 possible actions in experiments, thus 9 vectors); (b) objects coordinates, including the target position (4 objects, thus 4 vectors, one for each object class); (c) actor’s position in past steps (5 past steps, thus 5 vectors); (d) 1 feature plane for the walls in environments; and (e) 1 vector for the actor’s current position.

\paragraph{Prediction Net} Following spatialisation-concatenation operation, tensors are passed through a deep learning architecture.
The architectures utilised to conduct the reported experiments are formed of a shared prediction net torso and subsequently of separate prediction heads.
\begin{itemize}
\item The \textbf{torso} is implemented as a 2-layer ResNet with 32 channels, leaky ReLU nonlinearities, and batch-norm. The heads have all a similar architecture but differ  in terms of output size and format.
\item \textbf{Target Location Prediction Head.} The output from the torso is inputted into a 1-layer Convnet with 32 channels and leaky RELU, another 1-layer Convnet with 16 channels and leaky RELU, followed separately by (a) a fully connected layer to 121-dim logits (11 x 11 grid world) and (b) another 1-layer Convnet with 4 channels to 1. These are then summed, thus following a residual network approach. 
\item \textbf{Next Action Prediction Head}. The output from the torso is inputted into a 1-layer Convnet with 32 channels and leaky RELU, followed by average pooling, and 2 fully connected layers to 9-dimensions (9 possible actions).
\item \textbf{Next State Prediction Head}. The output from the torso is inputted into a 1-layer Convnet with 32 channels and leaky RELU, another 1-layer Convnet with 16 channels and leaky RELU, followed separately by (a) a fully connected layer to 121-dim logits (11 x 11 grid world) and (b) another 1-layer Convnet with 4 channels to 1. These are then summed.
\item \textbf{Beliefs Prediction Head}. The output from the torso is inputted into a 1-layer Convnet with 32 channels and leaky RELU, another 1-layer Convnet with 16 channels and leaky RELU, followed separately by (a) a fully connected layer to 121-dim logits (11 x 11 grid world) and (b) another 1-layer Convnet with 4 channels to 1. These are then summed.
\end{itemize}

\subsubsection{Agent Control System}
The agent has partial observability over the target position in the environment, i.e., they could see it only when it was in their 5x5 field of view, but they were fully informed of their position and that of the walls. To account for the resulting uncertainty and related information-gathering behaviours \cite{friston2015active}, the actors’ trajectories were generated using the POMDP planner based on Montecarlo tree search proposed by \cite{ognibene2019proactive} which extends \cite{silver2010monte}. It integrates a Bayesian filter that explicitly represents the actor’s \textit{beliefs} about the state of the task, i.e. the probability distribution on the target position. 
Note that while the beliefs design of the observed actor are necessarily determined by its task, this does not affect the generality of the observer’s performance because it is blind to the task and belief design assumptions.

\subsection{Environments}
\begin{figure}[thpb]
      \centering
      \framebox{
      \includegraphics[width=3in]{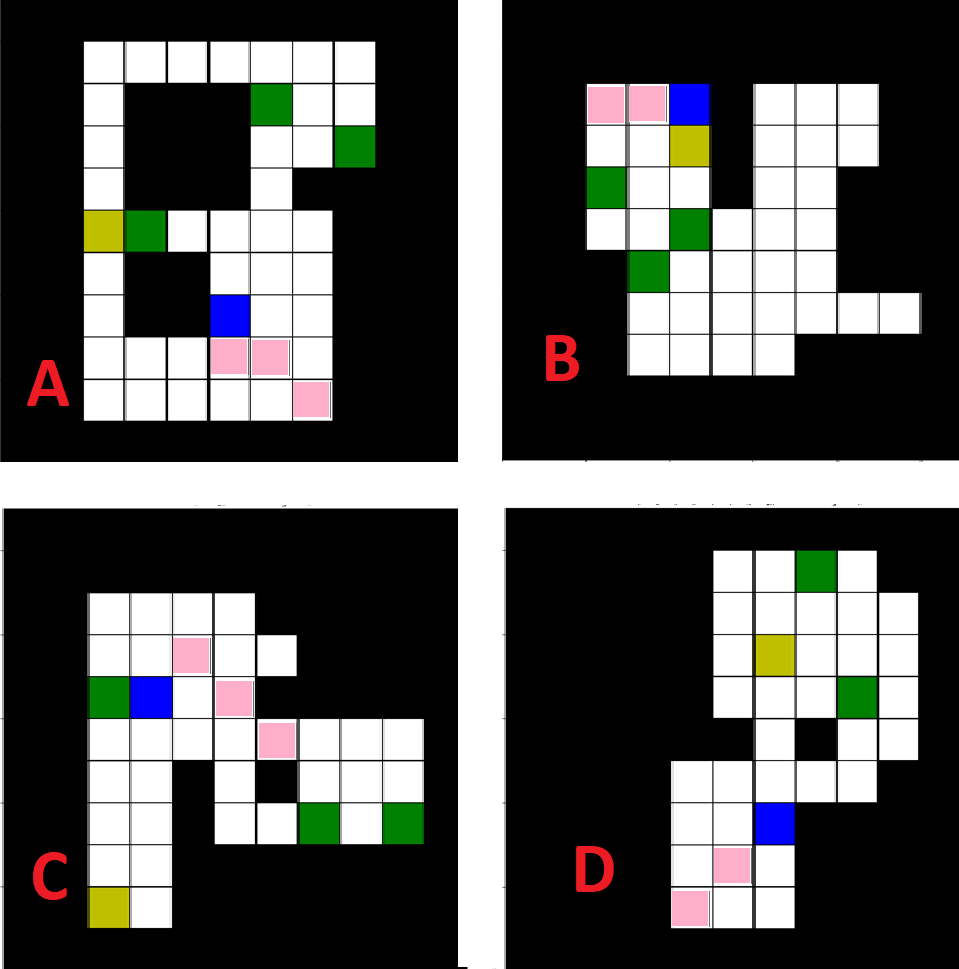}}
      \caption{Visualisation of example 11x11 grid world maps and observed trajectories, which varied in the location of walls, columns and free cells to move around the map. Colour code - Black: walls; White: empty cells; Yellow: target; Green: distractor objects; Blue: current actor’s position; Pink: past actor's positions.}
      \label{fig:maps}
   \end{figure}

For all experiments in this study, the environments consisted of $11 \times 11$ grid world maps, which varied in the location of walls, columns, and free cells to move around the map (see Fig. \ref{fig:maps} for a visualisation of example random maps and connected behaviours).  To assess the impact of dataset size on the proposed approach, we created multiple training datasets comprising  5, 10, 15, 20, 25, 30, 60, 120, and 300 maps, together with a 10 maps testing set.  The environments  enabled a common action space (north, east, south, west, northeast, northwest, southeast, southwest, stay) with deterministic results. During each trial, both training and testing,  the target, the distractors, and the agent are randomly positioned on free cells.

\subsection{Trajectory Dataset Generation}
A total of 30 trajectories were generated per grid world map by randomly selecting for each trajectory the initial locations of both the actor and target. For example, when considering the dataset with 60 grid world maps, a total of 1800 actors’ behaviours were created, while 9000 total behaviours were created for that with 300 maps. At training and test time, 3 distractor objects, identical to the target, were positioned at different empty locations on the map. The positioning of the distractor objects was random in experiments where it is not specified differently. This enabled a wider environment and behavioural diversity with no additional computational cost.

\subsection{Training}
All architectures (both Beliefs and NoBeliefs) were trained with the Adam optimiser, with varying learning rates (we tested 6 levels from 0.00015 to 0.001), using batches of size 32. A learning rate scheduler with the following parameters was used in all experiments: milestones = [30, 60, 80, 160], gamma = 0.5. The Cross Entropy loss function was utilised for all heads training, except for the Belief head, for which the Kullback–Leibler divergence loss function was used instead. This choice was driven by the fact that the Action, State and Target prediction heads all required one-hot encoding to identify one single position in the map, whereas a distribution of probabilities over each map location was needed for the Belief prediction head. A preliminary L1 and L2 regularisation tuning was performed using the L1 and L2 factor search; the final values of the L1 and L2 parameters used in all experiments were 0.005 and 0.001, respectively. In addition, early stopping was also integrated during training as a means to prevent overfitting. 

\section{RESULTS}
\subsection{Global comparison}
Table \ref{tbl:glbl_perf} shows the performance of the two architectures evaluated over the whole test set. The worst observed performance is $59.26$ (BEL architecture, 5 training maps) while the best is $72.43$ (BEL, 300 maps). The Beliefs architecture shows better performance than the NoBeliefs one in all conditions but the first one with 5  maps. The gain is statistically significant ($p-value<0.01$)  after the agent could observe the task performed over more than 20 training  maps (600 task executions). Under 5 training maps (150 observed task executions) the performance variance across trained nets is high. The additional learning constraint posed by the beliefs head seems to strengthen the over-fit of the network. However, over the learning trajectory, with a more reasonable dataset size, the beliefs prediction contributes to achieving higher performance.   The maximum performance difference observed is of  $1.89\%$ over a performance of the NoBeliefs architecture of $67.57\%$ when the agent was trained with 25 maps (750 task executions). Similar results were also obtained when changing the number of layers in the SPS components or the number of objects in the scene. 
An interesting trend to note is that the contribution of the beliefs attribution learning is stronger in the middle of the trajectory, when the architecture stops memorizing the beliefs samples and obtains  regularization effects, while becomes less useful with a high number of samples when regularization is not needed.  On the other side, the lower performance with 5 maps may point to the necessity of a complementary intention prediction dedicated system for such early stages, relating to the complexity of TOM development  observed in infants \cite{sodian2016theory,bianco2019transferring}.

\input{tables/glbltbl.tex}
\subsection{Comparison with hidden target}
Beliefs representations determine behavioural chunks when the target object is not still visible and its position is not deterministically known. In such conditions, shown in table \ref{tbl:glbl_perf_hidden_target}, the performance difference between the two architectures were stronger, with a maximum difference of about $7\%$ over a performance of the NoBeliefs architecture  $51\%$ (thus a relative gain of $14\%$) when the agent was trained with 15 maps (450 task executions). We must note also that this setting is also more challenging with the worst performance at $37.44\%$ and the best at $63.11\%$. Indeed, being the target position is not known to the agent, it should produce wandering behaviours that are not informative about the actual target position. However, the performance was above chance (25\%), we thus explored how the SPS could extract information from the presence of non-target objects, i.e. distractors.
\input{tables/hiddentbl.tex}
\subsection{Extracting information from  avoided distractors}
We reasoned that the most informative distractors would be those that were avoided (passed through) by the agent during his previous steps. Table \ref{tbl:neglect} shows the corresponding performance with a varying number of avoided, or better neglected, objects, with target visible or out of the field of view. The performance were much higher with $99\%$ for 3 avoided objects and visible target independently of the architecture and number of training maps. The worst performance were still at $79.67\%$ for 1 only avoided object and not visible target. Learning to predict beliefs is useful even when 3 objects have  been avoided, especially after few training maps. This implies that the gradient for the belief prediction  helps learning to discard the avoided objects. The maximum gain of the Beliefs architecture is $6.56$  is observed with 1 avoided objects and a high number of training maps. We assume that this reflects a strong contribution of the belief learning in discarding not yet avoided objects but that were not efficiently approached even when the agent does not still have goal approaching behaviour as it is still searching for the target. We moved next to study this specific condition.
\input{tables/neglecttbl.tex}
\subsection{Extracting information from  target-aligned distractors}
We assumed that the least informative distractors would be those that are aligned with future trajecotry of the agent toward the target. As table \ref{tbl:align} shows, the presence of such distractors strongly impacted the system performance with a worse performance of $11.56\%$ for 25 training maps, NoBeliefs architecture and 3 aligned objects. The gain of the Beliefs architecture where even stronger in this hard condition with a top gain of $12.83\%$ over a NoBeliefs performance of $43.83\%$, thus a relative gain of about $30\%$ during target visible codnitions. We interpreted this as a clue that the beliefs learning provides a more efficient way of learning to discard distractors that  have become visible before the targets even if they are aligned with it and not efficiently approached before. When the target is not visible the performance gain of Belief network is even stronger,  moving from a minimum go $27\%$ to a maximum of $37.28\%$. This shows that learning to predict beliefs helps learning to interpret beliefs driven searching behaviour that would be suboptimal in terms of object approaching. 
%
\input{tables/alignmenttbl.tex}
\subsection{Generalization to different agents}
Next we explored if learning to attribute beliefs would help interpret agents with a different decision system than the observer, and thus interpret  behaviours with a different style and effectiveness compared to those that were used for training under the self-observation hypothesis. We tested different ways, e.g. changing the action speed or beliefs representation precision, to differentiate the agents with consistent results. The results  shown in table \ref{tbl:general} were obtained decreasing the number of samples used by the POMCP algorithm \cite{silver2010monte}, which resulted in target search times  up to 4 times longer, and show a similar gain from learning to attribute beliefs as the data on observing similar agents, reported in table \ref{tbl:glbl_perf}. 
\input{tables/generalizationtbl.tex}


\section{CONCLUSIONS}
%
Beliefs play an important role in driving efficient behaviours when  the sensor system cannot capture instantaneously the whole state of the environment. 
Such  partial observable conditions are common in unstructured and natural environments in which artificial agents' ability to collaborate and support  humans is still limited. 
Dealing with the impact of partial observability and belief-driven behaviours would especially improve out-of-the-lab human robot interactions  and ego-vision-based systems \cite{damen2020epic}. 
Our results show the importance of the synergy of learning beliefs and intention attribution simultaneously, especially when limited  samples are available and acquiring new ones is expensive or risky. 
We showed that this architectural solution is particularly helpful in the most complex situations resulting from interpreting behaviours performed with limited sensory input, such as searching when multiple elements are in the field of view of the observed actor. 
We have also shown that the acquired TOM model, trained by the observing agent's own behaviours and mental processes (self-observation), is able to generalize to different agents while preserving the gain brought by the beliefs-intentions simultaneous learning. Future work will focus on extending these results to more complex and realistic settings to further test the applicability of this approach.

Finally, we believe that our findings on the importance of the learning synergy between  beliefs attribution and intention prediction  as well as the effectiveness of using own decision and belief estimation processes as learning signals to develop TOM  are  relevant beyond the development of social robot architectures \cite{rossi2017user,bianco2019,LEMAIGNAN201745,demiris2007prediction} and can also inform  the developmental psychology and neuroscience fields \cite{sodian2016theory,bianco2019transferring,brooks2015connecting,carr2018minding,peterson2019longitudinal,poulin2020testing}.







\addtolength{\textheight}{-12cm}   









\bibliographystyle{IEEEtran}
\bibliography{IEEEabrv,icra2023}

\end{document}

%% file: tables/glbltbl.tex
\begin{table}[b]
\caption{Target prediction accuracy}
\centering
\label{tbl:glbl_perf}
{\setlength{\tabcolsep}{4.5pt}
\tiny
\begin{tabular}{@{}c
>{\columncolor[HTML]{FCE4D6}}c 
>{\columncolor[HTML]{FCE4D6}}c 
>{\columncolor[HTML]{FCE4D6}}c 
>{\columncolor[HTML]{E2EFDA}}c 
>{\columncolor[HTML]{E2EFDA}}c 
>{\columncolor[HTML]{E2EFDA}}c 
>{\columncolor[HTML]{FFF2CC}}c 
>{\columncolor[HTML]{FFF2CC}}c @{}}
 & \multicolumn{3}{c}{\cellcolor[HTML]{F4B084}BEL} & \multicolumn{3}{c}{\cellcolor[HTML]{A9D08E}NoBEL} & \cellcolor[HTML]{FFD966} & \cellcolor[HTML]{FFD966} \\
\cellcolor[HTML]{F2F2F2}Train Maps (N) & \cellcolor[HTML]{F4B084}Best LR & \cellcolor[HTML]{F4B084}\begin{tabular}[c]{@{}c@{}}Avg Acc\\    (\%)\end{tabular} & \cellcolor[HTML]{F4B084}Var & \cellcolor[HTML]{A9D08E}Best LR & \cellcolor[HTML]{A9D08E}\begin{tabular}[c]{@{}c@{}}Avg Acc\\(\%)\end{tabular} & \cellcolor[HTML]{A9D08E}Var & \multirow{-2}{*}{\cellcolor[HTML]{FFD966}\begin{tabular}[c]{@{}c@{}}Bel-NoBel\\(\%)\end{tabular}} & \multirow{-2}{*}{\cellcolor[HTML]{FFD966}p-value} \\
\cellcolor[HTML]{F2F2F2}5 & 0.001 & 59.26 & 18.84 & 0.001 & 61.51 & 20.30 & \textbf{-2.25} & \textbf{.043} \\
\cellcolor[HTML]{F2F2F2}10 & 0.001 & 65.75 & 8.99 & 0.001 & 65.46 & 11.82 & 0.29 & .488 \\
\cellcolor[HTML]{F2F2F2}15 & 0.001 & 67.54 & 2.79 & 0.001 & 66.85 & 1.84 & 0.69 & .077 \\
\cellcolor[HTML]{F2F2F2}20 & 0.001 & 68.74 & 2.08 & 0.001 & 67.47 & 1.51 & \textbf{1.27} & \textbf{.001} \\
\cellcolor[HTML]{F2F2F2}25 & 0.001 & 69.45 & 1.63 & 0.001 & 67.57 & 1.44 & \textbf{1.87} & \textbf{\textless{}.001} \\
\cellcolor[HTML]{F2F2F2}30 & 0.001 & 69.49 & 1.60 & 0.001 & 67.79 & 1.76 & \textbf{1.69} & \textbf{\textless{}.001} \\
\cellcolor[HTML]{F2F2F2}60 & 0.00075 & 70.49 & 1.23 & 0.001 & 69.32 & 1.33 & \textbf{1.17} & \textbf{\textless{}.001} \\
\cellcolor[HTML]{F2F2F2}120 & 0.00075 & 71.47 & 0.87 & 0.0005 & 70.46 & 0.89 & \textbf{1.00} & \textbf{\textless{}.001} \\
\cellcolor[HTML]{F2F2F2}300 & 0.00015 & 72.43 & 0.28 & 0.00015 & 71.59 & 0.43 & \textbf{0.84} & \textbf{\textless{}.001}
\end{tabular}}
\vspace{6pt}

{Best target prediction accuracies by best learning rate obtained for the Beliefs vs NoBeliefs architectures based on the number of maps made available during training. Accuracies were calculated as averages across 18 initial network weights; the associated variances were reported.\\
BEL= Beliefs architecture; NoBEL = NoBeliefs architecture}
\end{table}

%% file: tables/hiddentbl.tex
\begin{table}[t]
\caption{Target prediction accuracy with hidden target and 3 distractors in view}
\label{tbl:glbl_perf_hidden_target}
{
\tiny
\begin{center}
\begin{tabular}{@{}c
>{\columncolor[HTML]{FCE4D6}}c 
>{\columncolor[HTML]{FCE4D6}}c 
>{\columncolor[HTML]{E2EFDA}}c 
>{\columncolor[HTML]{E2EFDA}}c 
>{\columncolor[HTML]{FFF2CC}}c 
>{\columncolor[HTML]{FFF2CC}}c @{}}
 & \multicolumn{5}{c}{\cellcolor[HTML]{B4C6E7}Target not   visible – 3 Objects visible} & \cellcolor[HTML]{B4C6E7} \\
 & \multicolumn{2}{c}{\cellcolor[HTML]{F4B084}BEL} & \multicolumn{2}{c}{\cellcolor[HTML]{A9D08E}NoBEL} & \cellcolor[HTML]{FFD966} & \cellcolor[HTML]{FFD966} \\
\cellcolor[HTML]{F2F2F2}\begin{tabular}[c]{@{}c@{}}Train Maps   \\    (N)\end{tabular} & \cellcolor[HTML]{F4B084}\begin{tabular}[c]{@{}c@{}}Avg Acc\\    (\%)\end{tabular} & \cellcolor[HTML]{F4B084}Var & \cellcolor[HTML]{A9D08E}\begin{tabular}[c]{@{}c@{}}Avg Acc\\    (\%)\end{tabular} & \cellcolor[HTML]{A9D08E}Var & \multirow{-2}{*}{\cellcolor[HTML]{FFD966}\begin{tabular}[c]{@{}c@{}}Bel-NoBel\\ (\%)\end{tabular}} & \multirow{-2}{*}{\cellcolor[HTML]{FFD966}p-value} \\
\cellcolor[HTML]{F2F2F2}5 & 37.44 & 96.97 & 38.56 & 31.44 & -1.11 & .681 \\
\cellcolor[HTML]{F2F2F2}10 & 52.61 & 29.31 & 48.83 & 17.44 & 3.78 & .025 \\
\cellcolor[HTML]{F2F2F2}15 & 58.17 & 27.09 & 51.11 & 16.69 & \textbf{7.06} & \textbf{\textless{}.001} \\
\cellcolor[HTML]{F2F2F2}20 & 56.94 & 8.29 & 54.17 & 12.85 & \textbf{2.78} & \textbf{.015} \\
\cellcolor[HTML]{F2F2F2}25 & 57.11 & 8.58 & 51.56 & 13.91 & \textbf{5.56} & \textbf{\textless{}.001} \\
\cellcolor[HTML]{F2F2F2}30 & 56.22 & 8.18 & 53.22 & 6.42 & \textbf{3.00} & \textbf{.002} \\
\cellcolor[HTML]{F2F2F2}60 & 59.39 & 19.08 & 54.44 & 12.73 & \textbf{4.94} & \textbf{.001} \\
\cellcolor[HTML]{F2F2F2}120 & 59.39 & 13.08 & 55.11 & 10.46 & \textbf{4.28} & \textbf{.001} \\
\cellcolor[HTML]{F2F2F2}300 & 61.61 & 5.19 & 63.11 & 4.34 & -1.50 & \textit{\textbf{.047}}
\end{tabular} 
\end{center}}

\vspace{6pt}
Best target prediction accuracies by best learning rate obtained for the Beliefs vs NoBeliefs architectures based on the number of maps made available during training. Accuracies were calculated as averages across 18 initial network weights; the associated variances were reported.
BEL= Beliefs architecture; NoBEL = NoBeliefs architecture
\end{table}

%% file: tables/neglecttbl.tex
\begin{table}[t]
\caption{Target prediction accuracies with varying number of avoided objects and target visibility}
\label{tbl:neglect}
{\tiny
\setlength{\tabcolsep}{2pt}
\begin{tabular}{@{}c
>{\columncolor[HTML]{FCE4D6}}c 
>{\columncolor[HTML]{FCE4D6}}c 
>{\columncolor[HTML]{E2EFDA}}c 
>{\columncolor[HTML]{E2EFDA}}c cc
>{\columncolor[HTML]{FCE4D6}}c 
>{\columncolor[HTML]{FCE4D6}}c 
>{\columncolor[HTML]{E2EFDA}}c 
>{\columncolor[HTML]{E2EFDA}}c cc@{}}
\multicolumn{1}{l}{} & \multicolumn{4}{c}{\cellcolor[HTML]{FFC000}Target visible – 25 maps} & \multicolumn{1}{l}{} & \multicolumn{1}{l}{} & \multicolumn{4}{c}{\cellcolor[HTML]{FFC000}Target not visible – 25 maps} & \multicolumn{1}{l}{} & \multicolumn{1}{l}{} \\
\multicolumn{1}{l}{} & \multicolumn{2}{c}{\cellcolor[HTML]{F4B084}BEL} & \multicolumn{2}{c}{\cellcolor[HTML]{A9D08E}NoBEL} & \multicolumn{1}{l}{} & \multicolumn{1}{l}{} & \multicolumn{2}{c}{\cellcolor[HTML]{F4B084}BEL} & \multicolumn{2}{c}{\cellcolor[HTML]{A9D08E}NoBEL} & \multicolumn{1}{l}{} & \multicolumn{1}{l}{} \\
\cellcolor[HTML]{E7E6E6}\begin{tabular}[c]{@{}c@{}} Neglected \\  (N)\end{tabular} & \cellcolor[HTML]{F4B084}\begin{tabular}[c]{@{}c@{}}Avg Acc\\  (\%)\end{tabular} & \cellcolor[HTML]{F4B084}Var & \cellcolor[HTML]{A9D08E}\begin{tabular}[c]{@{}c@{}}Avg Acc\\  (\%)\end{tabular} & \cellcolor[HTML]{A9D08E}Var & \cellcolor[HTML]{FFC000}Bel-NoBel & \cellcolor[HTML]{FFC000}\textit{p-val} & \cellcolor[HTML]{F4B084}\begin{tabular}[c]{@{}c@{}}Avg Acc\\  (\%)\end{tabular} & \cellcolor[HTML]{F4B084}Var & \cellcolor[HTML]{A9D08E}\begin{tabular}[c]{@{}c@{}}Avg Acc\\  (\%)\end{tabular} & \cellcolor[HTML]{A9D08E}Var & \cellcolor[HTML]{FFC000}Bel-NoBel & \cellcolor[HTML]{FFC000}\textit{p-val} \\
\cellcolor[HTML]{E7E6E6}3 & 99 & 0 & 99 & 0 & \cellcolor[HTML]{FFF2CC}0 & \cellcolor[HTML]{FFF2CC}1 & 98.56 & 0.51 & 97.33 & 0.84 & \cellcolor[HTML]{FFF2CC}1.22 & \cellcolor[HTML]{FFF2CC}{\color[HTML]{FF0000} \textless{}.001} \\
\cellcolor[HTML]{E7E6E6}2 & 98.89 & 0.1 & 98.44 & 0.26 & \cellcolor[HTML]{FFF2CC}0.44 & \cellcolor[HTML]{FFF2CC}{\color[HTML]{FF0000} 0.01} & 94.17 & 0.38 & 91.78 & 2.07 & \cellcolor[HTML]{FFF2CC}2.39 & \cellcolor[HTML]{FFF2CC}{\color[HTML]{FF0000} \textless{}.001} \\
\cellcolor[HTML]{E7E6E6}1 & 97.72 & 0.21 & 96.67 & 0.59 & \cellcolor[HTML]{FFF2CC}1.06 & \cellcolor[HTML]{FFF2CC}{\color[HTML]{FF0000} \textless{}.001} & 85.28 & 0.68 & 81.89 & 6.46 & \cellcolor[HTML]{FFF2CC}3.39 & \cellcolor[HTML]{FFF2CC}{\color[HTML]{FF0000} \textless{}.001} \\
\multicolumn{1}{l}{} & \multicolumn{4}{c}{\cellcolor[HTML]{FFC000}Target visible – 120 maps} & \multicolumn{1}{l}{} & \multicolumn{1}{l}{} & \multicolumn{4}{c}{\cellcolor[HTML]{FFC000}Target not visible – 120 maps} & \multicolumn{1}{l}{} & \multicolumn{1}{l}{} \\
\cellcolor[HTML]{E7E6E6}3 & 99 & 0 & 99 & 0 & \cellcolor[HTML]{FFF2CC}0 & \cellcolor[HTML]{FFF2CC}1 & 98.94 & 0.24 & 98.72 & 0.46 & \cellcolor[HTML]{FFF2CC}0.22 & \cellcolor[HTML]{FFF2CC}0.77 \\
\cellcolor[HTML]{E7E6E6}2 & 99 & 0 & 98.61 & 0.37 & \cellcolor[HTML]{FFF2CC}0.39 & \cellcolor[HTML]{FFF2CC}{\color[HTML]{FF0000} 0.01} & 94.67 & 0.24 & 90.78 & 13.48 & \cellcolor[HTML]{FFF2CC}3.89 & \cellcolor[HTML]{FFF2CC}{\color[HTML]{FF0000} \textless{}.001} \\
\cellcolor[HTML]{E7E6E6}1 & 98.44 & 0.26 & 96.83 & 1.44 & \cellcolor[HTML]{FFF2CC}1.61 & \cellcolor[HTML]{FFF2CC}{\color[HTML]{FF0000} \textless{}.001} & 86.22 & 0.77 & 79.67 & 33.65 & \cellcolor[HTML]{FFF2CC}6.56 & \cellcolor[HTML]{FFF2CC}{\color[HTML]{FF0000} \textless{}.001}
\end{tabular}}
\vspace{6pt}

{Target prediction accuracies for the Beliefs vs NoBeliefs architectures in the conditions with varying number of neglected objects and target visible by the actor (25 maps (A), 120 maps (C)) or target not visible by the actor (25 maps (B), 120 maps (C)). Accuracies were calculated as averages across 18 initial network weights; the associated variances were reported.}
\end{table}

%% file: tables/alignmenttbl.tex
\begin{table}[t]
\caption{Target prediction accuracies with varying number of aligned objects and target visibility}
\label{tbl:align}
{\tiny
\setlength{\tabcolsep}{1.5pt}
\begin{tabular}{c
>{\columncolor[HTML]{FCE4D6}}c 
>{\columncolor[HTML]{FCE4D6}}c 
>{\columncolor[HTML]{E2EFDA}}c 
>{\columncolor[HTML]{E2EFDA}}c cc
>{\columncolor[HTML]{FCE4D6}}c 
>{\columncolor[HTML]{FCE4D6}}c 
>{\columncolor[HTML]{E2EFDA}}c 
>{\columncolor[HTML]{E2EFDA}}c cc}
 & \multicolumn{4}{c}{\cellcolor[HTML]{FFC000}Target visible - 25 maps} &  &  & \multicolumn{4}{c}{\cellcolor[HTML]{FFC000}Target not visible - 25 maps} &  &  \\
 & \multicolumn{2}{c}{\cellcolor[HTML]{F4B084}BEL} & \multicolumn{2}{c}{\cellcolor[HTML]{A9D08E}NoBEL} &  &  & \multicolumn{2}{c}{\cellcolor[HTML]{F4B084}BEL} & \multicolumn{2}{c}{\cellcolor[HTML]{A9D08E}NoBEL} &  &  \\
\cellcolor[HTML]{E7E6E6}\begin{tabular}[c]{@{}c@{}}N Objects\\  Aligned\end{tabular} & \cellcolor[HTML]{F4B084}\begin{tabular}[c]{@{}c@{}}Avg Acc\\  (\%)\end{tabular} & \cellcolor[HTML]{F4B084}Var & \cellcolor[HTML]{A9D08E}\begin{tabular}[c]{@{}c@{}}Avg Acc\\  (\%)\end{tabular} & \cellcolor[HTML]{A9D08E}Var & \cellcolor[HTML]{FFC000}Bel-NoBel & \cellcolor[HTML]{FFC000}\textit{p-val} & \cellcolor[HTML]{F4B084}\begin{tabular}[c]{@{}c@{}}Avg Acc\\  (\%)\end{tabular} & \cellcolor[HTML]{F4B084}Var & \cellcolor[HTML]{A9D08E}\begin{tabular}[c]{@{}c@{}}Avg Acc\\  (\%)\end{tabular} & \cellcolor[HTML]{A9D08E}Var & \cellcolor[HTML]{FFC000}Bel-NoBel & \cellcolor[HTML]{FFC000}\textit{p-val} \\
\cellcolor[HTML]{E7E6E6}3 & 43.5 & 12.85 & 36.22 & 12.65 & \cellcolor[HTML]{FFF2CC}7.28 & \cellcolor[HTML]{FFF2CC}{\color[HTML]{FF0000} \textless{}.001} & 16.61 & 14.02 & 11.56 & 8.38 & \cellcolor[HTML]{FFF2CC}5.06 & \cellcolor[HTML]{FFF2CC}{\color[HTML]{FF0000} \textless{}.001} \\
\cellcolor[HTML]{E7E6E6}2 & 40.94 & 19.7 & 31.61 & 15.78 & \cellcolor[HTML]{FFF2CC}9.33 & \cellcolor[HTML]{FFF2CC}{\color[HTML]{FF0000} \textless{}.001} & 17.56 & 23.91 & 10.83 & 10.5 & \cellcolor[HTML]{FFF2CC}6.72 & \cellcolor[HTML]{FFF2CC}{\color[HTML]{FF0000} \textless{}.001} \\
\cellcolor[HTML]{E7E6E6}1 & 51.44 & 18.97 & 45.78 & 28.07 & \cellcolor[HTML]{FFF2CC}5.67 & \cellcolor[HTML]{FFF2CC}{\color[HTML]{FF0000} 0.001} & 42.56 & 33.08 & 34.83 & 37.44 & \cellcolor[HTML]{FFF2CC}7.72 & \cellcolor[HTML]{FFF2CC}{\color[HTML]{FF0000} \textless{}.001} \\
 & \multicolumn{4}{c}{\cellcolor[HTML]{FFC000}Target visible - 120 maps} &  &  & \multicolumn{4}{c}{\cellcolor[HTML]{FFC000}Target not visible - 120 maps} &  &  \\
\cellcolor[HTML]{E7E6E6}\begin{tabular}[c]{@{}c@{}}N Objects \\ Aligned\end{tabular} & \cellcolor[HTML]{F4B084}\begin{tabular}[c]{@{}c@{}}Avg Acc\\  (\%)\end{tabular} & \cellcolor[HTML]{F4B084}Var & \cellcolor[HTML]{A9D08E}\begin{tabular}[c]{@{}c@{}}Avg Acc\\  (\%)\end{tabular} & \cellcolor[HTML]{A9D08E}Var & \cellcolor[HTML]{FFC000}Bel-NoBel & \cellcolor[HTML]{FFC000}\textit{p-val} & \cellcolor[HTML]{F4B084}\begin{tabular}[c]{@{}c@{}}Avg Acc\\  (\%)\end{tabular} & \cellcolor[HTML]{F4B084}Var & \cellcolor[HTML]{A9D08E}\begin{tabular}[c]{@{}c@{}}Avg Acc\\  (\%)\end{tabular} & \cellcolor[HTML]{A9D08E}Var & \cellcolor[HTML]{FFC000}Bel-NoBel & \cellcolor[HTML]{FFC000}\textit{p-val} \\
\cellcolor[HTML]{E7E6E6}3 & 49.33 & 6.82 & 43.22 & 14.18 & \cellcolor[HTML]{FFF2CC}6.11 & \cellcolor[HTML]{FFF2CC}{\color[HTML]{FF0000} \textless{}.001} & 15.39 & 16.02 & 12.11 & 16.34 & \cellcolor[HTML]{FFF2CC}3.28 & \cellcolor[HTML]{FFF2CC}{\color[HTML]{FF0000} 0.02} \\
\cellcolor[HTML]{E7E6E6}2 & 46.61 & 3.01 & 36.83 & 4.05 & \cellcolor[HTML]{FFF2CC}9.78 & \cellcolor[HTML]{FFF2CC}{\color[HTML]{FF0000} \textless{}.001} & 14.94 & 20.17 & 10.89 & 16.22 & \cellcolor[HTML]{FFF2CC}4.06 & \cellcolor[HTML]{FFF2CC}{\color[HTML]{FF0000} 0.007} \\
\cellcolor[HTML]{E7E6E6}1 & 56.67 & 15.88 & 43.83 & 12.97 & \cellcolor[HTML]{FFF2CC}12.83 & \cellcolor[HTML]{FFF2CC}{\color[HTML]{FF0000} \textless{}.001} & 46.28 & 33.27 & 35.06 & 31.35 & \cellcolor[HTML]{FFF2CC}11.22 & \cellcolor[HTML]{FFF2CC}{\color[HTML]{FF0000} \textless{}.001}
\end{tabular}}
\vspace{6pt}

{Target prediction accuracies for the Beliefs vs NoBeliefs architectures in the conditions with varying number of aligned objects and target visibility. (A) Target visible, 25 maps; (B) Target not visible, 25 maps; (C) Target visible, 120 maps; and (D) Target not visible, 120 maps. Accuracies were calculated as averages across 18 initial network weights; the associated variances were reported}
\end{table}

%% file: tables/generalizationtbl.tex
\begin{table}[t]%
\caption{Target prediction accuracies with varying observed agent efficiency}%
\label{tbl:general}%
{\setlength{\tabcolsep}{3pt}%
\begin{center}%
\begin{tabular}{c
>{\columncolor[HTML]{FCE4D6}}c 
>{\columncolor[HTML]{FCE4D6}}c 
>{\columncolor[HTML]{E2EFDA}}c 
>{\columncolor[HTML]{E2EFDA}}c 
>{\columncolor[HTML]{FFF2CC}}c 
>{\columncolor[HTML]{FFF2CC}}c }
\multicolumn{7}{c}{\cellcolor[HTML]{FFC000}\textit{\textbf{250 max samples – Original}}} \\
\multicolumn{1}{l}{} & \multicolumn{2}{c}{\cellcolor[HTML]{F4B084}\textit{\textbf{Bel}}} & \multicolumn{2}{c}{\cellcolor[HTML]{A9D08E}\textit{\textbf{NoBel}}} & \cellcolor[HTML]{FFD966} & \cellcolor[HTML]{FFD966} \\
\cellcolor[HTML]{F2F2F2}\textit{\textbf{\begin{tabular}[c]{@{}c@{}}Train Maps \\  (N)\end{tabular}}} & \cellcolor[HTML]{F4B084}\textit{\textbf{\begin{tabular}[c]{@{}c@{}}Avg Acc\\  (\%)\end{tabular}}} & \cellcolor[HTML]{F4B084}\textit{\textbf{Var}} & \cellcolor[HTML]{A9D08E}\textit{\textbf{\begin{tabular}[c]{@{}c@{}}Avg Acc\\  (\%)\end{tabular}}} & \cellcolor[HTML]{A9D08E}\textit{\textbf{Var}} & \multirow{-2}{*}{\cellcolor[HTML]{FFD966}\textit{\textbf{\begin{tabular}[c]{@{}c@{}}Bel-NoBel\\  (\%)\end{tabular}}}} & \multirow{-2}{*}{\cellcolor[HTML]{FFD966}\textit{\textbf{p-val}}} \\
\cellcolor[HTML]{F2F2F2}\textit{\textbf{25}} & \textit{\textbf{69.45}} & \textit{\textbf{1.63}} & \textit{\textbf{67.57}} & \textit{\textbf{1.44}} & \textit{\textbf{1.87}} & {\color[HTML]{FF0000} \textit{\textbf{\textless{}.001}}} \\
\cellcolor[HTML]{F2F2F2}\textit{\textbf{120}} & \textit{\textbf{71.47}} & \textit{\textbf{0.87}} & \textit{\textbf{70.46}} & \textit{\textbf{0.89}} & \textit{\textbf{1}} & {\color[HTML]{FF0000} \textit{\textbf{\textless{}.001}}} \\
\multicolumn{7}{c}{\cellcolor[HTML]{FFC000}\textit{\textbf{150 max samples}}} \\
\cellcolor[HTML]{F2F2F2}\textit{\textbf{25}} & \textit{\textbf{68.61}} & \textit{\textbf{2.25}} & \textit{\textbf{67.44}} & \textit{\textbf{2.73}} & \textit{\textbf{1.17}} & {\color[HTML]{FF0000} \textit{\textbf{0.033}}} \\
\cellcolor[HTML]{F2F2F2}\textit{\textbf{120}} & \textit{\textbf{70.94}} & \textit{\textbf{1.47}} & \textit{\textbf{70.39}} & \textit{\textbf{0.96}} & \textit{\textbf{0.56}} & \textit{\textbf{0.139}} \\
\multicolumn{7}{c}{\cellcolor[HTML]{FFC000}\textit{\textbf{50 max samples}}} \\
\cellcolor[HTML]{F2F2F2}\textit{\textbf{25}} & \textit{\textbf{68.22}} & \textit{\textbf{1.59}} & \textit{\textbf{66.78}} & \textit{\textbf{1.59}} & \textit{\textbf{1.44}} & {\color[HTML]{FF0000} \textit{\textbf{0.002}}} \\
\cellcolor[HTML]{F2F2F2}\textit{\textbf{120}} & \textit{\textbf{70.11}} & \textit{\textbf{1.28}} & \textit{\textbf{69.61}} & \textit{\textbf{0.96}} & \textit{\textbf{0.5}} & \textit{\textbf{0.165}} \\
\multicolumn{7}{c}{\cellcolor[HTML]{FFC000}\textit{\textbf{25 max samples}}} \\
\cellcolor[HTML]{F2F2F2}\textit{\textbf{25}} & \textit{\textbf{65.83}} & \textit{\textbf{1.21}} & \textit{\textbf{65}} & \textit{\textbf{1.18}} & \textit{\textbf{0.83}} & {\color[HTML]{FF0000} \textit{\textbf{0.028}}} \\
\cellcolor[HTML]{F2F2F2}\textit{\textbf{120}} & \textit{\textbf{68.5}} & \textit{\textbf{0.74}} & \textit{\textbf{68.61}} & \textit{\textbf{1.31}} & {\color[HTML]{FF0000} \textit{\textbf{-0.11}}} & \textit{\textbf{0.744}} \\
\end{tabular}
\end{center}}
\vspace{6pt}

{Target prediction accuracies for the Beliefs vs NoBeliefs architectures in the conditions with varying complexity of the actor’s cognitive capabilities. Number of max samples underlying actor’s behaviour: (A) 250 (Original); (B) 150; (C) 50; (D) 25.
Accuracies were calculated as averages across 18 initial network weights; the associated variances were reported.}
\end{table}